\documentclass{OAGM}
\OAGMarXiv{1304.1876}

\usepackage{graphicx}
\usepackage{tabularx}
\usepackage[compact,noindentafter]{titlesec}
\titlespacing\section{0pt}{5pt plus 4pt minus 4pt}{5pt plus 2pt minus 2pt}
\titlespacing\subsection{0pt}{5pt plus 4pt minus 4pt}{5pt plus 2pt minus 2pt}
\titlespacing\subsubsection{0pt}{5pt plus 4pt minus 4pt}{5pt plus 2pt minus 2pt}

\title{A Bag of Visual Words Approach for Symbols-Based Coarse-Grained Ancient Coin Classification}

\author{Hafeez Anwar, Sebastian Zambanini \and Martin Kampel\\
  Computer Vision Lab, Vienna University of Technology, Vienna, Austria}

\begin{document}
\maketitle

\begin{abstract}
The field of Numismatics provides the names and descriptions of the symbols minted on the ancient coins. Classification of the ancient coins aims at assigning a given coin to its issuer. Various issuers used various symbols for their coins. We propose to use these symbols for a framework that will coarsely classify the ancient coins. Bag of visual words (BoVWs) is a well established visual recognition technique applied to various problems in computer vision like object and scene recognition. Improvements have been made by incorporating the spatial information to this technique. We apply the BoVWs technique to our problem and use three symbols for coarse-grained classification. We use rectangular tiling, log-polar tiling and circular tiling to incorporate spatial information to BoVWs. Experimental results show that the circular tiling proves superior to the rest of the methods for our problem.
\end{abstract}

\section{Introduction}
%\vspace{-10pt}
In this paper, we address the problem of visual classification of ancient coins. Our proposed visual classification framework is based on the symbols minted on the reverse side of the coins. These symbols have specific names and descriptions for which we have to refer to \emph{Numismatics} \cite{Numis} which deals with the study of coins and currency, specifically the branch of Numismatics concerned with the knowledge of ancient coins. These ancient coins are indexed in standard reference books \cite{Crawford74}. The heads of a coin is called the \textbf{obverse side} while the tails is called the \textbf{reverse side}. For Roman Republican coins, which are the focus of this work, the obverse side usually depicts a portrait of an emperor or a god while the reverse side has embossed drawings of instruments, weapons, animals and temples etc \cite{Crawford74}. We call these embossed drawings \emph{symbols} and use them for the classification of ancient coins. These symbols serve as a natural and intuitive choice for classification. The descriptions to an ancient coin given by an expert source include the name of a king whose portrait is on the obverse side, the year of his reign and the description of the symbols on the reverse side. Therefore besides the Latin inscriptions, the symbols on the reverse side provide the most natural cue to classify a coin.\newline
The manual classification of coins from sources like private and museum collections or new coin findings requires time and expertise. Therefore an image-based coin classification framework can provide an improved and faster classification that can be conducted by non-experts, as manual coin classification can be done only by experts due to its complexity \cite{Numis}.

\subsection*{Research issues}
The visual classification of ancient coins comes with its own set of research challenges. The available dataset for this research contains about 3900 Roman Republican coins that belong to 550 distinct classes. By class we mean the visually distinguishable coin types defined in the standard reference book \cite{Crawford74}. There is a significant variability in the coins of the same class whereas coins of different classes possibly have significant visual similarity. The wear and tear of a coin due to its age as well as the conditions in which it was preserved lead to local variations within a class and missing of certain visually important parts. Apart from this, due to lack of technology in ancient times, the coins were minted by different die engravers with use of different dies. Therefore various parts of a coin of the same class are not necessarily identical.

\begin{figure}[t]
  \centering
  % Requires \usepackage{graphicx}
  \includegraphics[width=100mm, height=80mm,keepaspectratio]{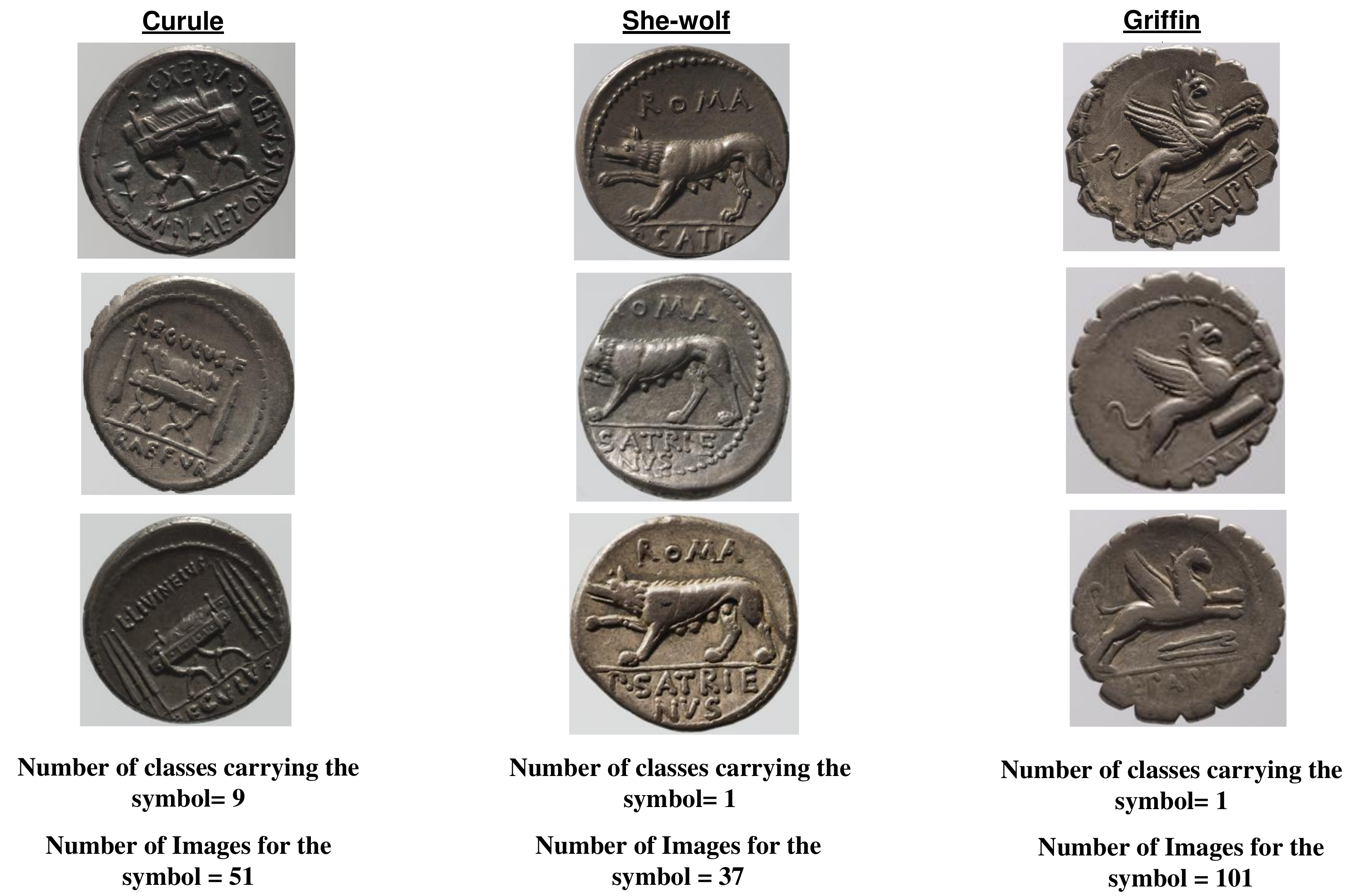}
  \caption{Symbols used for classification, number of classes and images for each symbol}\label{fig1}
\end{figure}

%\begin{table}[htb]
%    \setlength{\belowcaptionskip}{-12pt}
%    \setlength{\abovecaptionskip}{-2pt}
%  \begin{center}
%
%\begin{tabularx}{1\textwidth}{ |>{\setlength\hsize{1\hsize}\centering}X|>{\setlength\hsize{1\hsize}\centering}X|>{\setlength\hsize{1\hsize}\centering}X| }
%  \hline
%Name of Symbol & {Number of classes containing the symbol} & Number images of symbol\tabularnewline
%\hline
%  Griffin & 1 & 101 \tabularnewline
%  \hline
%  Curule & 9 & 51 \tabularnewline
%  \hline
%  She wolf & 1 & 37\tabularnewline
%  \hline
%\end{tabularx}
%
%\end{center}
%  \caption{Number of visual examples and classes in dataset for each symbol}
%  \label{tab1}
%\end{table}

\subsection*{Coarse-grained classification}
In this paper we aim at a coarse-grained classification of ancient Roman coins as a given symbol may be minted on coins of more than one class. We consider three symbols named griffin, curule and she-wolf as shown in Fig.~\ref{fig1}. The pronounced variance in curule can be seen in Fig.~\ref{fig1} as there are 9 classes in the dataset on which it is minted. We will categorize the coins based on these three symbols. Fig.~\ref{fig1} also shows the number of classes in our dataset on which these symbols are minted along with the number of visual examples for each symbol.
\subsection*{Related work}
There is little work on application of computer vision techniques for the visual classification of ancient coins. Most of the work focuses on the recognition of modern coins. However modern day coins are manufactured using sophisticated technology as compared to the dies used in ancient times. Apart from that, the effects on modern coins due to wear and tear are almost negligible as compared to ancient coins. The proposed techniques for modern day coins classification based on gradient information \cite{Nölle03,Maaten} and Eigenspaces \cite{Huber:2005} are not sufficient for ancient coins classification as shown by Zaharieva et al. \cite{Zaharieva:2007}. SIFT features \cite{Lowe:2004} are used by Kampel and Zaharieva \cite{Kampel08} to obtain the classification accuracy of ~90\% on a dataset of 390 images belonging to only 3 classes. Sparse SIFT features are matched between images and the ones with the greatest number of matches are assigned to the same class. The method proposed by Arandjelovi\'{c} \cite{Aran2010} combines geometry with SIFT by calculating the directional histograms at the positions of the keypoints. Although the method proved superior to BoVWs with a classification accuracy of 57.2\% against 2.4\% yet the BoVWs representation used lack the spatial information. Additionally, experiments were only performed on the obverse sides of the coins belonging to 65 classes. The method proposed in \cite{Zamba12} is based on a dense correspondence search between coin images. The visual similarity of coins is derived from these correspondences and exploited for a coarse-to-fine search for the coin exemplar in the database with highest similarity. The method achieves a classification rate  of 82.8\% on a test set containing 60 classes of Roman Republican coins.\newline
We propose the first method which explicitly achieves symbol recognition on coins. We propose to incorporate more knowledge from Numismatics than any other method for ancient coins classification. This method can be extended to a comprehensive classification framework which will use semantically meaningful information from expert sources of Numismatics. The central idea of our approach is very close to human intuition as we rely on the symbols minted on the reverse side. These symbols have sufficient differences in their structures due to which they are more discriminating than the portraits of the emperors on the obverse side. The contributions of this paper are as follows:
\vspace{-3pt}
\begin{itemize}
  \setlength{\belowcaptionskip}{-12pt}
  \setlength{\abovecaptionskip}{-5pt}
  \setlength{\itemsep}{-2pt}
  \item Coarse-grained classification of ancient coins based on symbols.
  \item The use of dense SIFT based BoVWs representation for ancient coins classification.
  \item Evaluation of three approaches (rectangular, circular and log-polar), for capturing the spatial information of BoVWs for the task of ancient coins classification.
\end{itemize}
\vspace{-3pt}
The rest of the paper is organized as follows: Section 2 gives an overview of the bag of visual words on which our proposed framework is based. The details of our proposed framework are given in Section 3. Experiments and results are reported in Section 4. Finally we conclude the paper in Section 5 and give an overview of the future directions for research.

\section{Bag of Visual Words (BoVWs)}
\begin{figure}[t]
  \centering
  % Requires \usepackage{graphicx}
  \includegraphics[width=120mm, height=100mm,keepaspectratio]{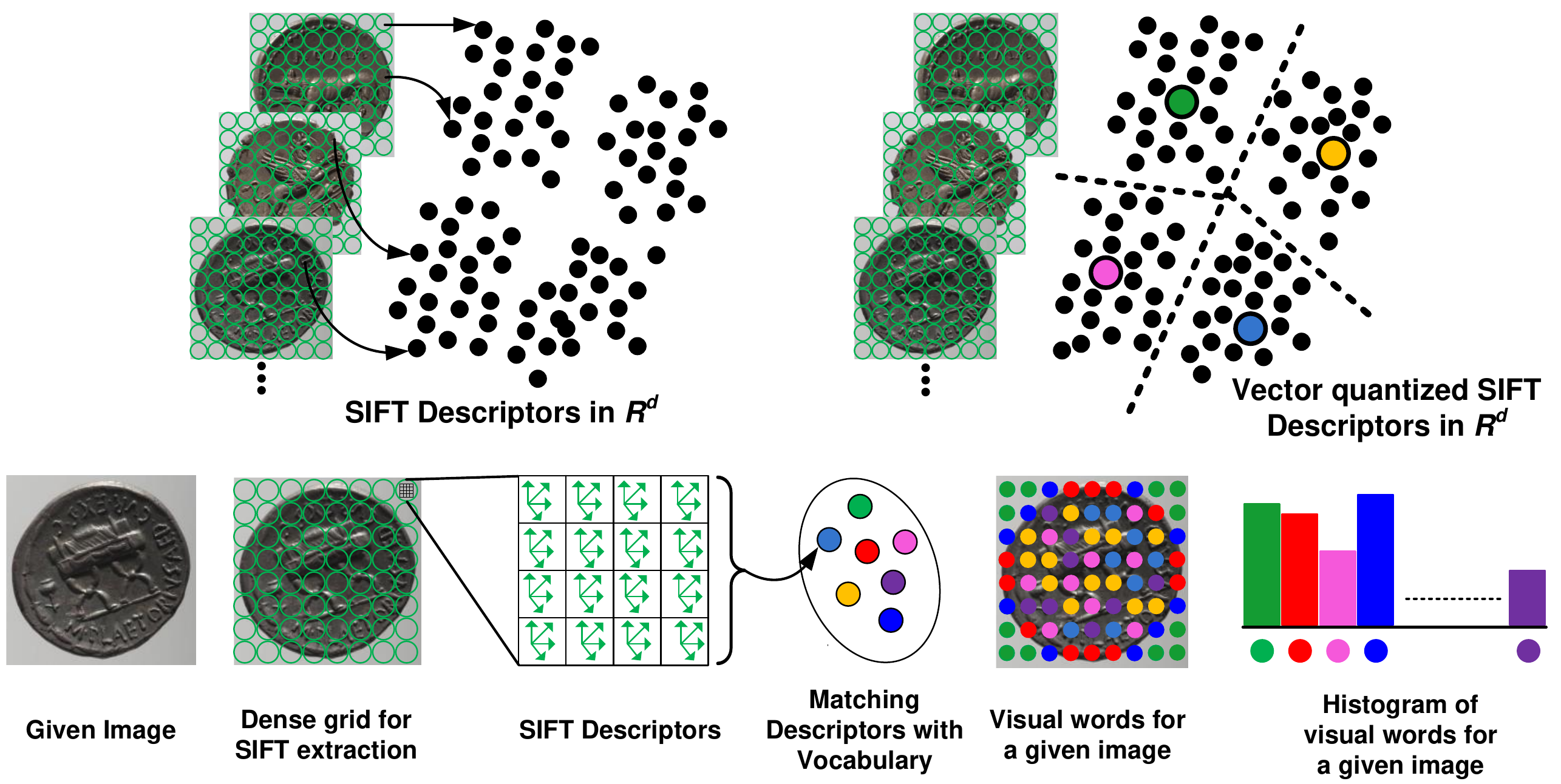}
  \caption{Procedure for vocabulary generation and BoVWs image representation}\label{fig2}
\end{figure}

%\vspace{-10pt}
The BoVWs method of image representation draws its concept from text documents. A text document usually consists of a distribution of words. A compact representation of any textual document can be visualized by the distributions of words in it. Following the same concept, an image can be represented using ``visual words'' \cite{Sivic03}. The important visual information for the representation of an image is concentrated in its interest points or keypoints. Here we use Lowe's SIFT descriptors for image representation. We use a dense sampling approach which deals with the extraction of SIFT features on a regular grid across the image. However due to the high dimensionality of these descriptors, they are inappropriate to be used as visual words. Therefore a quantization method is imposed on the feature space of these image descriptors to build a visual vocabulary. Following steps are performed to build a visual vocabulary.
\vspace{-3pt}
\begin{enumerate}
  \setlength{\itemsep}{-2pt}
  \item Features are extracted from a representative image set.
  \item A quantization scheme is used to cluster the image features. For instance, k-means clustering is used where $k$ is the size of the visual vocabulary.
  \item The features of a novel image are then mapped according to the established visual vocabulary. The mapping is based on Euclidean distance between a word and a given descriptor.
  \item Finally the novel image is represented as a histogram of visual words.
\end{enumerate}
\vspace{-5pt}
Fig.~\ref{fig2} gives a visual summary of the procedure for BoVWs image representation.
\vspace{-10pt}

\section{Adding Spatial Information to BoVWs Representation}
\begin{figure}[h!tb]
  \centering
  % Requires \usepackage{graphicx}
  \includegraphics[width=120mm, height=100mm,keepaspectratio]{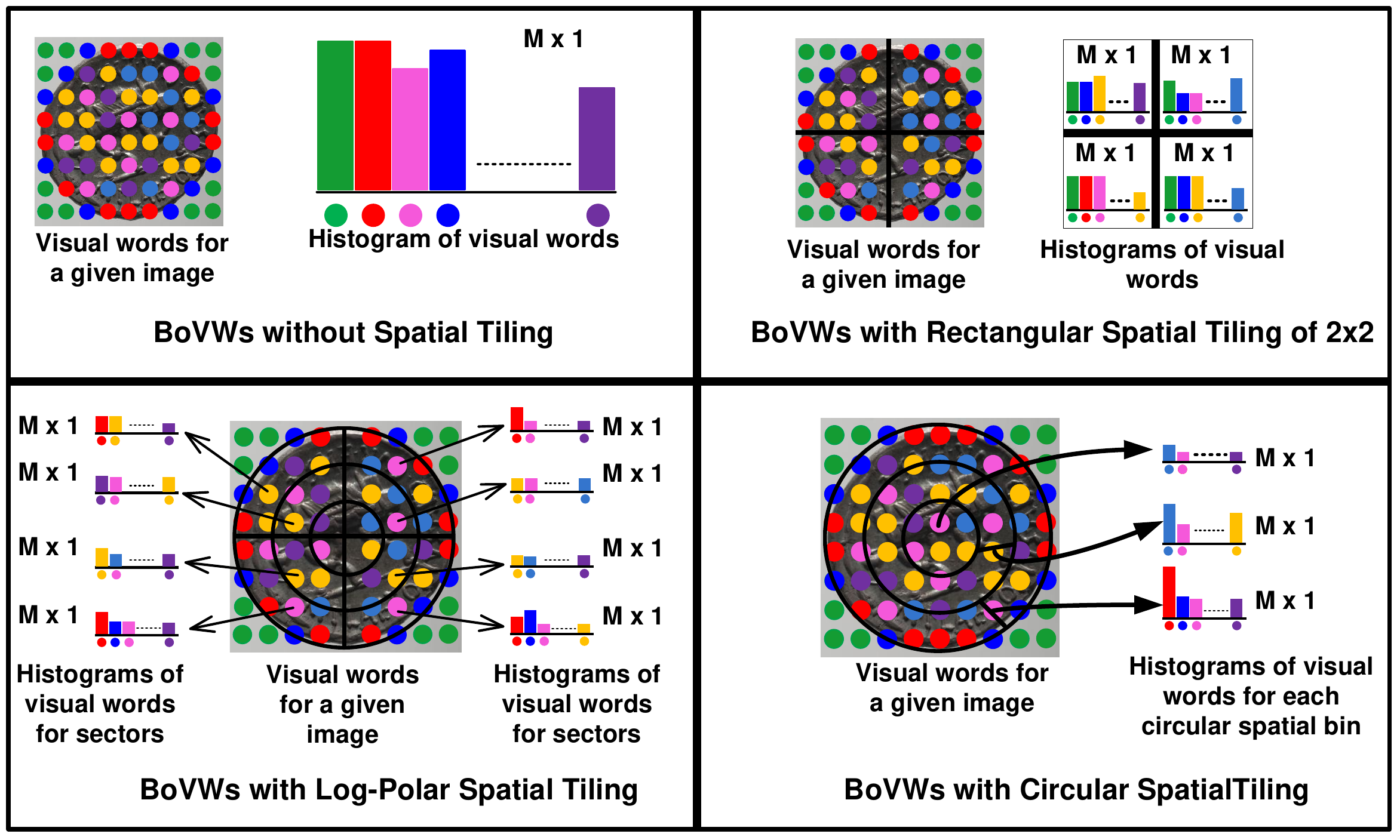}
  \caption{BoVWs image representation with and without spatial tilings}\label{fig3}
\end{figure}

There are issues of this technique that have to be investigated for our application scenario. The first issue is the size of vocabulary for BoVWs. A small vocabulary leads to an insufficient representation of all image patches while a large vocabulary causes quantization artifacts and overfitting. We thus evaluate a range of visual vocabularies for our task. The second and most important issue is the lack of geometry as the BoVWs approach does not take the spatial location of the visual words into account. Although this characteristic of BoVWs make them flexible to viewpoint and pose changes yet it can result in a low performance at places where spatial information is an important discriminating factor. Since we will use symbols for coarse-grained ancient coins classification, these symbols have specific geometric structures and their parts are spatially related to each other. Therefore incorporating the spatial information to BoVWs for our task will lead to significant improvements in performance and this is shown in the results. Here we give an overview of the methods for incorporating spatial information to BoVWs. The illustration of these methods is given in Fig.~\ref{fig3}.

\subsection{Rectangular Tiling}
For a given image, the densely sampled SIFT features are mapped to the visual words of a vocabulary of size $M$. The image is then partitioned into $2\times2$ regions. Therefore instead of representing the image by an $M$ sized histogram, it is represented by 4 histograms of size $M$ for each region. After computing the histograms for all image regions, they are concatenated into a single feature vector of size $M \times 4$.  However this method of spatial tiling is not invariant to rotations. If the image is rotated sufficiently, then the visual words of one image partition fall into the adjacent partition leading to a change in the histograms of all the four regions.

\subsection{Log-polar Tiling}
Belongie et al. \cite{Belongie01} used the log-polar binning scheme for the first time to develop a descriptor for shape matching. Recently, log-polar spatial tiling was used by Zhang and Mayo \cite{Zhang:2010} to incorporate spatial information to BoVWs. According to the results presented in their paper, the proposed spatial scheme outperformed the Spatial Pyramid Matching (SPM) of Lazebnik et al. \cite{Lazebnik:2006} at three benchmark datasets. By applying the log-polar spatial tiling, a given image is divided into sectors of various scales and orientations. This detailed spatial sectoring of an image leads to fine capturing of the distribution of image features both in distance and orientations. Log-polar spatial tiling is added to the BoVWs representation of an image by performing the following steps:
\vspace{-5pt}
\begin{itemize}
  \setlength{\itemsep}{-2pt}
  \item A log-polar tiling scheme of $r$ scales and $\theta$ orientations is imposed on the image resulting in a total of $r \times \theta$ spatial sectors of the image. In our approach, we empirically use $r = 3$ and $\theta= 4$.
  \item For each sector in the log-polar tiling, a histogram of visual words of size $M$ is calculated.
  \item Finally, the histograms of all the log-polar sectors are concatenated into a single feature vector of size $M \times r \times \theta$.
\end{itemize}

This approach is similar to the rectangular tiling but the difference lies in the way image subregions are defined. The size of the image subregions vary as their distance from the center of the image increases. As shown in Fig.~\ref{fig3}, the log-polar sectors near the center of the image are smaller than the ones far from the center. This results in a spatial tiling where the sectors far from the center of the image contain more number of visual words than those near to it. The detailed splitting of the image into log-polar sectors helps to capture the spatial relationship of the image patches but this scheme is also not invariant to rotations. The rotation of an image will lead to a shift of patches in the adjacent sectors and hence the distributions of the visual words will change in them. This change in distributions will affect the histograms of all the  sectors.
\subsection{Circular Tiling}
We propose to use circular tiling to incorporate the spatial information to BoVWs as shown in Fig.~\ref{fig3}. Circular tiling is a natural choice for our problem as ancient coins are circular, though not perfectly circular due to reasons mentioned in Section 1. We impose concentric circular tiling to generate the BoVWs representation. After the image is represented by the visual words from a vocabulary of size $M$, circular binned spatial information is added to it as follows.
\begin{itemize}
  \setlength{\itemsep}{-2pt}
  \item A concentric circular tiling is imposed on this representation consisting of $r$ circles. We examine $r = 3$ and use the smaller dimension among width and height of a given image to calculate the radii of circles so that the outer circle lies inside the image boundaries.
  \item For each circular bin, a histogram of visual words of size $M$ is calculated.
  \item Finally, the histograms of all the circular bins are concatenated into a single feature vector of size $M \times r$.

\end{itemize}
Circular tiling is invariant to rotations because if a coin is rotated, it will not result in change of the distributions of visual words in circular bins. Therefore the histograms based on circular tiling will remain invariant to rotation of the coin.
\section{Experiments and Results}
\begin{figure}[h!bt]
   \centering
  % Requires \usepackage{graphicx}
  \includegraphics[width=120mm, height=100mm,keepaspectratio]{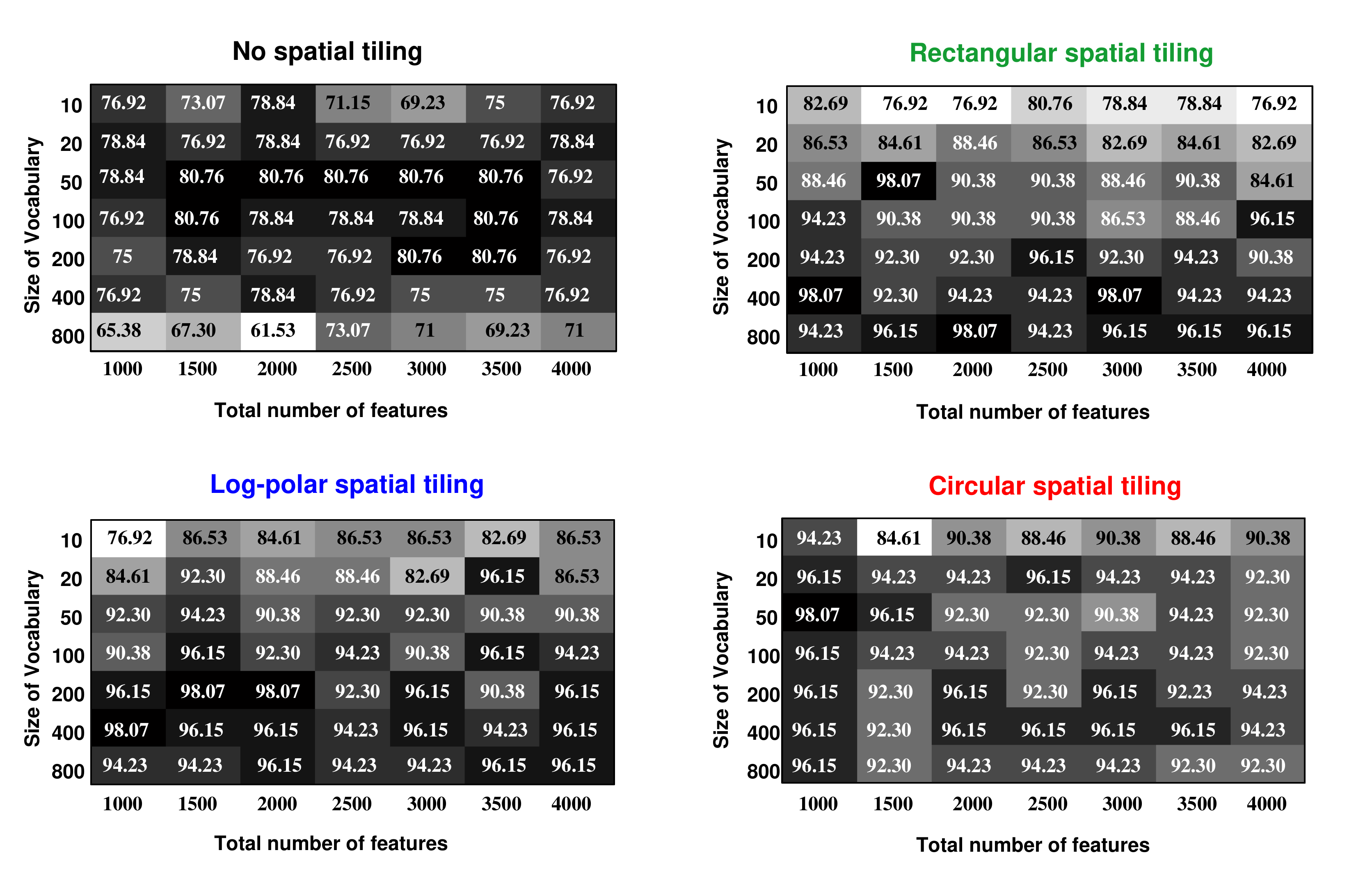}
  \caption{Classification accuracy as a function of vocabulary size and total number of features}\label{fig6}
\end{figure}

\begin{figure}[!h!bt]
  \centering
  % Requires \usepackage{graphicx}
  \includegraphics[width=80mm, height=70mm,keepaspectratio]{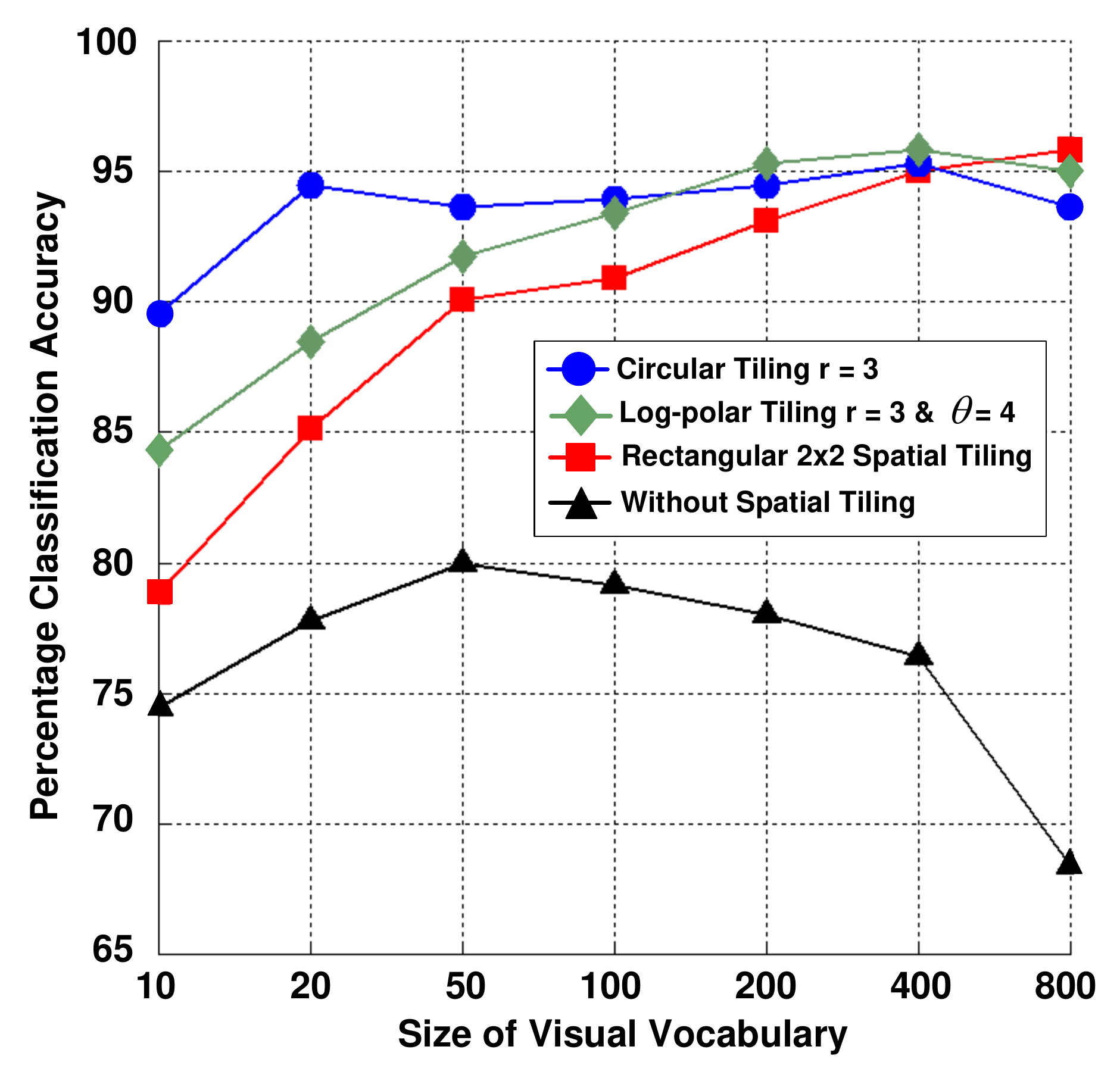}
  \caption{Mean accuracy for all number of features for all vocabulary sizes}\label{fig7}
\end{figure}

%\vspace{-10pt}
 As stated before, we consider three symbols of ancient coins for our experiments. We evaluate each spatial scheme for a given number of vocabulary sizes and total number of SIFT features used to build the vocabulary. During the training phase, dense SIFT features are extracted from each one of the training images. As the total number of features extracted from the training images can be quite large, a random subset of features is selected. The size of this random subset is varied as \{1000, 1500, 2000, 2500, 3000, 3500, 4000\}. However equal number of features are selected from all the training images. k-means clustering is then used to cluster the features in order to construct the vocabulary. Only the training set is used to construct the visual vocabulary. The vocabulary sizes used are \{10, 20, 50, 100, 200, 400, 800\}. 70\% of the images are used as training set and the remaining 30\% as test set. A one-vs-all multi-class classification is carried out using the SVM classifier of LIBSVM library \cite{libsvm}. We use an RBF kernel for our experiments where the best values for $C$ and $\gamma$ are determined with n-fold cross-validation on the training set. The classification accuracy for various combinations of size of vocabulary and number of features for rectangular, log-polar and circular tiling is shown in Fig.~\ref{fig6}.
To visualize the comparison of all the three methods collectively, for each vocabulary size, we took an average of the classification accuracies for all the features. Fig.~\ref{fig7} shows the comparison. It can be seen that all the three methods are superior to the BoVWs representation with no spatial information. The circular tiling outperforms rest of the spatial schemes on vocabulary sizes of 10, 20, 50 and 100 but as the vocabulary size increases, the behavior of all the three schemes converge. We claimed in the previous section that circular tiling is superior to the rest of the methods in terms of rotation invariance. We figured out that the converging nature of the three methods is due to the lack of a challenging rotation variance of the images in the test set. Symbols on the images in our test set are not too much rotated leading to a lack of difference in performance of the three methods. Therefore, circular tiling can be considered as the best choice for our problem as it provides rotation invariance without losing discriminative power compared to rectangular and log-polar tiling.

\section{Conclusion and Future Work}
We proposed a framework for the visual classification of ancient coins based on the symbols minted on their reverse side. Three symbols were considered for classification. BoVWs approach was used to represent the images. As BoVWs image representation lacks the spatial information, three types of spatial tilings were evaluated to incorporate the spatial information. Rectangular, log-polar and circular tiling were used and evaluated. All the three schemes showed superior performance to BoVWs representation where spatial information is not incorporated. However circular tiling outperformed rest of the two schemes on our dataset. In the future we plan to validate our promising results on a larger number of symbols and more variation of coin rotation in the dataset. Consequently, the framework will be extended towards a larger-scale system in order to allow for a fast and robust recognition of significant symbols on ancient coins.
\subsection*{Acknowledgments}
This research was supported by the Austrian Science Fund (FWF) under the grant TRP140-N23-2010 (ILAC) and the Vienna PhD School of Informatics (http://www.informatik.tuwien.ac.at/teaching/phdschool).

\bibliography{refs}

\begin{thebibliography}{10}

\bibitem{Aran2010}
O.~Arandjelovi{\'c}.
\newblock Automatic attribution of ancient {R}oman imperial coins.
\newblock In {\em CVPR}, pages 1728--1734, 2010.

\bibitem{Belongie01}
S.~Belongie, J.~Malik, and J.~Puzicha.
\newblock Shape matching and object recognition using shape contexts.
\newblock {\em TPAMI}, 24:509--522, 2001.

\bibitem{libsvm}
Chih-Chung Chang and Chih-Jen Lin.
\newblock {LIBSVM}: A library for support vector machines.
\newblock {\em ACM Transactions on Intelligent Systems and Technology},
  2:27:1--27:27, 2011.

\bibitem{Crawford74}
Michael~H. Crawford.
\newblock {\em Roman Republican Coinage, 2 vols}.
\newblock Cambridge University Press, 1974.

\bibitem{Numis}
P.~Grierson.
\newblock Numismatics.
\newblock Oxford University Press, 1975.

\bibitem{Huber:2005}
Reinhold H., Herbert R., Konrad M., Harald P., and Michael R.
\newblock Classification of coins using an eigenspace approach.
\newblock {\em Pattern Recogn. Lett.}, 26(1):61--75, January 2005.

\bibitem{Kampel08}
M.~Kampel and M.~Zaharieva.
\newblock Recognizing ancient coins based on local features.
\newblock In {\em ISVC}, volume~I of {\em LNCS}, pages 11--22. Springer-Verlag,
  2008.

\bibitem{Lazebnik:2006}
S.~Lazebnik, C.~Schmid, and J.~Ponce.
\newblock Beyond bags of features: spatial pyramid matching for recognizing
  natural scene categories.
\newblock In {\em CVPR}, pages 2169--2178, 2006.

\bibitem{Lowe:2004}
David~G. Lowe.
\newblock Distinctive image features from scale-invariant keypoints.
\newblock {\em Int. J. Comput. Vision}, 60:91--110, 2004.

\bibitem{Maaten}
L.~Van~Der Maaten and P.~Boon.
\newblock {COIN - O - MATIC: A fast system for reliable coin classification}.
\newblock In {\em MUSCLE CIS Coin Competition Workshop}, pages 07--18, 2006.

\bibitem{Nölle03}
M.~N\"{o}lle, H.~Penz, M.~Rubik, K.~Mayer, I.~Holl\"{a}nder, and R.~Granec.
\newblock Dagobert - a new coin recognition and sorting system.
\newblock In {\em DICTA}, pages 329--338, 2003.

\bibitem{Sivic03}
J.~Sivic and A.~Zisserman.
\newblock {Video Google}: {A} text retrieval approach to object matching in
  videos.
\newblock In {\em ICCV}, volume~2, pages 1470--1477, 2003.

\bibitem{Zaharieva:2007}
M.~Zaharieva, M.~Kampel, and S.~Zambanini.
\newblock Image based recognition of ancient coins.
\newblock In {\em CAIP}, pages 547--554, 2007.

\bibitem{Zamba12}
S.~Zambanini and M.~Kampel.
\newblock Coarse-to-fine correspondence search for classifying ancient coins.
\newblock In {\em 2nd ACCV Workshop on e-Heritage}, 2012.

\bibitem{Zhang:2010}
E.~Zhang and M.~Mayo.
\newblock Enhanced spatial pyramid matching using log-polar-based image
  subdivision and representation.
\newblock In {\em DICTA}, pages 208--213, 2010.

\end{thebibliography}
\end{document}